\def\year{2016}
\documentclass[letterpaper]{article}
\usepackage{aaai16,times,helvet,courier}
\usepackage{amsmath,amsfonts,amssymb,sectsty,graphicx,float}
\newcommand\addtag{\refstepcounter{equation}\tag{\theequation}}
\newcommand{\al}{\alpha}
\renewcommand{\b}[1]{\boldsymbol{#1}}
\newcommand{\bal}{\boldsymbol{\alpha}}
\newcommand{\bbe}{\boldsymbol{\beta}}

\newcommand{\be}{\beta}
\newcommand{\bet}{\boldsymbol{\eta}}
\newcommand{\bpi}{\boldsymbol{\pi}}
\newcommand{\bphi}{\boldsymbol{\phi}}
\newcommand{\bs}{\backslash}
\newcommand{\bth}{\boldsymbol{\theta}}
\newcommand{\by}{\boldsymbol{y}}
\newcommand{\bZ}{\boldsymbol{Z}}
\renewcommand{\c}{,\!}
\newcommand{\cM}{\mathcal{M}}
\newcommand{\de}{\delta}

\newcommand{\Ga}{\Gamma}
\newcommand{\ga}{\gamma}
\newcommand{\iid}{\stackrel{iid}{\sim}}
\newcommand{\ka}{\kappa}
\newcommand{\la}{\lambda}
\newcommand{\si}{\sigma}
\newcommand{\Si}{\Sigma}
\newcommand{\tbpi}{\widetilde{\boldsymbol{\pi}}}
\newcommand{\tG}{\widetilde{G}}
\renewcommand{\th}{\theta}
\newcommand{\tpi}{\widetilde{\pi}}
\newcommand{\vphi}{\varphi}
\begin{document}
\title{Temporal Topic Analysis with Endogenous and Exogenous Processes}
\author{Baiyang Wang, Diego Klabjan\\
Department of Industrial Engineering and Management Sciences,\\
Northwestern University, 2145 Sheridan Road, Evanston, Illinois, USA, 60208\\
baiyang@u.northwestern.edu\qquad d-klabjan@northwestern.edu}
\maketitle
\begin{abstract}
\begin{quote}
We consider the problem of modeling temporal textual data taking endogenous and exogenous processes into account. Such text documents arise in real world applications, including job advertisements and economic news articles, which are influenced by the fluctuations of the general economy. We propose a hierarchical Bayesian topic model which imposes a "group-correlated" hierarchical structure on the evolution of topics over time incorporating both processes, and show that this model can be estimated from Markov chain Monte Carlo sampling methods. We further demonstrate that this model captures the intrinsic relationships between the topic distribution and the time-dependent factors, and compare its performance with latent Dirichlet allocation (LDA) and two other related models. The model is applied to two collections of documents to illustrate its empirical performance: online job advertisements from DirectEmployers Association and journalists' postings on BusinessInsider.com.
\end{quote}
\end{abstract}
\section{Introduction}
Many organizations nowadays provide portals for job posting and job search, such as glassdoor.com from Glassdoor, indeed.com from Recruit, and my.jobs from DirectEmployers Association. Our work is inspired by data collected from the portal my.jobs, a website where job seekers can apply to the posted job openings through a provided link. The data collected from the website includes user clickstreams (users create accounts on the site) and attributes of job advertisements, such as their description, location, company name, and posted date.

In this paper, we investigate the relationship between economic fluctuations and the related changes in job advertisements, which can reveal the economic conditions of different time periods. More generally, this question is about the influence of any exogenous process on textual data with temporal dimensions. We adopt the perspective that the documents are organized into a certain number of topics, and study the impact of the exogenous process on the topic distribution, i.e. the relative topic proportions. Given a corpus of text documents with time stamps and a related exogenous process, the problem is to find a relationship between the topics discussed and the exogenous process. This setting is natural in an economic context; for instance, changes in macroeconomic indicators have an impact on government reports and {\it Wall Street Journal} news articles. Meanwhile, we also notice that for most temporal documents, the topic proportions change over time, which indicates an endogenous process of topic evolution. 

With the goal of establishing topic dependency on the endogenous and exogenous processes, LDA-type topic models are especially suitable. The latent Dirichlet allocation (LDA) \cite{B03} is the original model. Since then, a large number of variants have been proposed, many of which can be found in Blei (2011). 

Meanwhile, there has been relatively limited discussion on modeling time-dependent documents when there are relevant simultaneous exogenous processes. Many time-dependent topic models without the exogenous component have been proposed, such as the Topics over Time (ToT) model \cite{W06} and the dynamic topic model (DTM) \cite{B06}, to name a few. However, to the best of our knowledge, none of these papers incorporate the effect of exogenous processes. On the other hand, the structural topic model (STM) \cite{R15} considers the effect of metadata, i.e. the attributes specified for each document, on the topic distribution. While STM can be applied for mining time-dependent textual data with exogenous covariates, it does not explicitly consider the time factor or the endogenous topic evolution processes of time-stamped documents.

Our approach to this problem is to incorporate both endogenous and exogenous processes into a topic model. For the endogenous part of our paper, we impose a Markovian structure on the topic distribution over time, similar to Blei and Lafferty (2006) and Dubey et al. (2014). For the exogenous process, we incorporate it into the topic distribution in each period, adjusting the endogenous topic evolution process. In this way, our model is essentially a stick-breaking truncation of a ''group-correlated'' hierarchical Dirichlet process. Our model has the following contributions: (i) it addresses the question of measuring the influence of exogenous processes on the topics in related documents, (ii) it incorporates both endogenous and exogenous aspects, and (iii) it demonstrates that text mining can also have useful implications in the realm of economics, which, from the authors' perspective, is a relatively new finding.

Section 2 offers a brief review on the topic modeling techniques related to our model. Section 3 develops our hierarchical Bayesian model and describes how to make posterior inferences with a variant of the Markov chain Monte Carlo (MCMC) technique. Section 4 studies the online job advertisements from DirectEmployers Association and journalists' postings in finance on BusinessInsider.com with our proposed method, providing a comparison of performance with the standard LDA and STM. Section 5 suggests possible directions for the future and concludes the paper.

\section{Review of Time-Dependent Topic Modeling}
We first introduce the standard model of LDA \cite{B03}. Suppose that there is a collection of documents $d_i$, $i=1,\ldots,N$ and words $\{x_{i,j}\}_{j=1}^{J_i}$ within each document $d_i$ indexed by a common dictionary containing $V$ words, where $N$ is the number of documents, and $J_i$ is the number of words in $d_i$. The LDA model is as follows,
\[
\begin{cases}
\bth_i\iid Dir(\bal),\ \bphi_k \iid Dir(\bbe), \\
z_{i,j}|\bth_i\iid Cat(\bth_i),\  x_{i,j}|z_{i,j}\sim Cat(\bphi_{z_{i,j}}).\\
\end{cases}\addtag
\]
Here $i=1,\ldots,N$, $j=1,\ldots,J_i$, $k=1,\ldots,K$; $\bth_i$ is the length-$K$ per-document topic distribution for $d_i$, $\bphi_k$ is the length-$V$ per-topic word distribution for the $k$-th topic, $z_i^j$ is the topic for the $j$-th word in $d_i$, and $K$ is the number of topics. $Dir(\cdot)$ denotes the Dirichlet distribution and $Cat(\cdot)$ denotes the categorical distribution, a special case of the multinomial distribution when $n_{obs}=1$. 

The Dirichlet process is a class of randomized probability measures and can be applied for non-parametric modeling of mixture models. Denoting the concentration parameter by $\al$ and the mean probability measure by $H$, a realization $G$ from the Dirichlet process can be written as $G\sim DP(\ga, H)$. With the stick-breaking notation \cite{S94}, we have
\begin{equation}
G = \sum_{k=1}^\infty b_k\de_{\vphi_k},
\end{equation}
where $\de_{\vphi_k}$ is a ``delta'' probability measure with all the probability mass placed at $\vphi_k$, $\vphi_k\iid H$, $b_k=b_k'\prod_{i=1}^{k-1}(1-b_i')$, $b_k'\iid Beta(1,\ga)$, $k=1,2,\cdots$. We write $\b{b} = (b_1, b_2, \ldots) \sim Stick(\ga)$. More properties of the Dirichlet process can be found in Ferguson (1973).

A hierarchical Dirichlet process (HDP) was proposed in the context of text modeling by Teh et al. (2005). The following hierarchical structure is assumed,
\[
\begin{cases}
G_0|\ga\sim DP(\ga, H), \\
G_1, \ldots, G_N|(\al, G)\iid DP(\al, G), \\
\bphi_{i,j}|G_i\iid G_i,\ x_{i,j}\sim Cat(\bphi_{i,j}).
\end{cases}\addtag
\]
Here $i=1,\ldots,N$, $j=1,\ldots,J_i$. The length-$V$ random vectors $G_0, G_1, \ldots, G_N$ are ``random word distributions,'' each of which is a draw from a Dirichlet process in (3). Moreover, each draw from a random word distribution is a length-$V$ fixed vector $\bphi_{i,j}$; it is the word distribution for $x_{i,j}$. The posterior inference can be achieved by different strategies of Gibbs sampling.

There are mainly two approaches in the literature of measuring endogenous topic evolution processes. One approach is to impose a finite mixture structure on the topic distribution: a dynamic hierarchical Dirichlet process (dHDP) \cite{R08} was proposed by adding a temporal dimension, and its variation was further applied on topic modeling with a stick-breaking truncation of Dirichlet processes \cite{P10}. The other approach imposes a Markovian structure. For instance, the dynamic topic model (DTM) \cite{B06} is as follows,
\[
\begin{cases}
\bphi_{t,k}|\bphi_{t-1,k}\sim N(\bphi_{t-1,k}, \si^2I),\\
\bal_t|\bal_{t-1}\sim N(\bal_{t-1}, \de^2I),\\
\bth_{t,i}|\bal_t\iid N(\bal_t, a^2I),\\
z_{t,i,j}|\bth_{t,i}\iid Cat(\exp(\bth_{t,i})),\\
x_{t,i,j}|z_{t,i,j}\sim Cat(\exp(\bphi_{t,z_{t,i,j}})).
\end{cases}\addtag
\]
Here $t=1, \ldots, T$, $i=1,\ldots,N_t$, $j=1,\ldots,J_{t,i}$, $k=1,\ldots,K$ ($t\ge2$ for the first two equations); $T$ is the number of time periods, $N_t$ is the number of documents in the $t$-th period, and $J_{t,i}$ is the number of words in the $i$-th document in the $t$-th period; the rest are similarly defined as in LDA. One major difference between DTM and LDA is that the topic distributions $\bth_{t,i}$ and word distributions $\bphi_{t,k}$ are in log-scale in DTM. A variational Kalman filtering was proposed for the posterior inference. As this Markovian approach is simpler for both interpretation and posterior inference, we apply a more generalized version of it to specify the endogenous process in our model.

The structural topic model (STM) \cite{R15} measures the effect of metadata of each document with the logistic normal distribution. Their model for each document $d_i$ is as follows,
\[
\begin{cases}
\bth_i|(X_i\ga,\Si)\sim LogisticNormal(X_i\ga,\Si),\\ p(\bphi_{i,k})\propto\exp(m+\ka_k+\ka_{g_i}+\ka_{kg_i}),\\
z_{i,j}|\bth_i\iid Cat(\bth_i),\  x_{i,j}|z_{i,j}\sim Cat(\bphi_{i,z_{i,j}}),
\end{cases}\addtag
\]
where $i=1,\ldots,N$, $j=1,\ldots,J_i$; $X_i$ is the metadata matrix, $\ga$ is a coefficient vector, $\Si$ is the covariance matrix, $\bphi_{i,k}$ is the word distribution for $d_i$ and the $k$-th topic, $m$ is a baseline log-word distribution, $\ka_k$, and $\ka_{g_i}$ and $\ka_{kg_i}$ are the topic, group, and interaction effects; the rest are defined similarly to LDA. This model explicitly considers exogenous factors, and can be applied to find the relationship between topic distributions and exogenous processes. Below we adopt a slightly more general approach, incorporating both endogenous and exogenous factors.

\section{Model and Algorithm}
\subsection{Motivation: A Group-Correlated Hierarchical Dirichlet Process}
We formulate our problem as follows: we are given time periods $t = 1,\ldots,T$, documents from each period $d_{t,i}$, $i=1,\ldots,N_t$, $t=1,\ldots,T$, and the indices of words $\{x_{t,i,j}\}_{j=1}^{J_{t,i}}$ within each document $d_{t,i}$ from the first word to the last. The words are indexed by a dictionary containing $V$ words in total. We begin with a hierarchical Dirichlet process in time $1$: let $G_1|\ga\sim DP(\ga, H)$, $G_{1i}|(\al_1, G_1)\sim DP(\al_1, G_1)$, where $G_1$ is a baseline random word distribution for time $1$, and $G_{1i}$ is the random word distribution for document $d_{1i}$. For $G_2,\ldots G_T$, we have the following Markovian structure,
\begin{equation}
p(G_t)|G_{t-1}\propto\exp[-d(G_t,G_{t-1})],\ t=2,\ldots,T.
\end{equation}
Here $d(\cdot,\cdot)$ is some distance between two probability measures. This completes our endogenous process. To take an exogenous process $\{\by_t\}_{t=1}^T$ into account, we assume the following
\begin{equation}
\tG_{t} = \cM(G_t, \by_t),\ t=1,\ldots,T,
\end{equation}
where $\cM$ maps the endogenous baseline random word distribution $G_t$ to the realized baseline random word distribution $\tG_t$ for time $t$, considering the influence of $\{\by_t\}_{t=1}^T$. Therefore, we further assume that each per-document random word distribution $G_{t,i}$ is sampled with mean $\tG_t$ rather than $G_t$. The final model is as follows,
\[
\begin{cases}
G_1|\ga\sim DP(\ga, H),\\
p(G_t)|G_{t-1}\propto\exp[-d(G_t,G_{t-1})],\\
\tG_{t} = \cM(G_t, \by_t),\\
G_{t,i}|(\al_t, \tG_t) \iid DP(\al_t, \tG_t),\\
\bphi_{t,i,j}|G_{t,i}\iid G_{t,i},\ x_{t,i,j}\sim Cat(\bphi_{t,i,j}).
\end{cases}\addtag
\]
Here $t = 1,\ldots,T$, $i=1,\ldots,N_t$, $j=1, \ldots, J_{t,i}$ ($t\ge2$ for the first line). Throughout this paper, our model is fully conditional on $\{\by_t\}_{t=1}^T$, i.e. we assume $\{\by_t\}_{t=1}^T$ to be fixed; this has an intuitive explanation, as our temporal documents represent a very small portion of the underlying environment, i.e. the exogenous process, so their influence on $\{\by_t\}_{t=1}^T$ is almost negligible. 
\subsection{A Group-Correlated Temporal Topic Model: Stick-Breaking Truncation}
Below we consider a stick-breaking truncation of the model above, since posterior inference of the exact model can be intricate. With the stick-breaking expression of $G_1$ in (8), we have
\[
\begin{cases}
\bphi_1, \bphi_2, \ldots, \iid H,\\
\bpi_1 = (\pi_{11},\pi_{12},\ldots)\sim Stick(\ga),\\
G_1 = \sum_{k=1}^\infty \pi_{1k}\de_{\bphi_k}.
\end{cases}\addtag
\]
Here we set $d(\cdot,\cdot)=+\infty$ if the two probability measures have different supports; this necessitates that all periods share the same topics. Our intent is that the topics should remain the same to investigate their relationships with endogenous and exogenous processes; otherwise, changes in topics can blur the relationships and possibly result in overfitting. We apply the total variation distance $d(p,q)=\la\cdot\int|p-q|d\mu$ with $\la>0$, although many others can also be applied and lead to, for instance, a log-normal model in DTM, or a normal model \cite{D14,Z15}. We have the following,
\[
\begin{cases}
G_t = \sum_{k=1}^\infty \pi_{tk}\de_{\bphi_k},\\
\pi_{tk} = \pi_{t-1\, k} + Lap(\la),\\
\bpi_t = (\pi_{t1},\pi_{t2},\ldots).
\end{cases}\addtag
\]
Here $t=2,\ldots,T$, $k=1,2,\ldots$, and $Lap(\la)$ denotes a Laplacian distribution with scale parameter $\la$. For the exogenous part, we consider specifying the relationship between $\bpi_t$ and $\tbpi_t = (\tpi_{t1},\tpi_{t2},\ldots)$ such that $\tG_t = \sum_{k=1}^\infty \tpi_{tk}\de_{\bphi_k}$, $G_{t,i}=\sum_{k=1}^\infty \theta_{t,i,k}\delta_{\bphi_k}$. We let 
\begin{equation}
\tbpi_t = \bpi_t + \bet\cdot\by_t,\ t=1,\ldots,T,\ \b1'\cdot\bet = \b0.
\end{equation}
Here $\bet$ is a $K\times p$ matrix which indicates the relationship between the topic distribution $\tbpi_t$ and the length-$p$ vector $\by_t$. However, we notice that $\tbpi_t$ and $\bpi_t$ are of infinite length, which creates difficulty in our inference. Therefore we adopt a stick-breaking truncation approach, i.e. we only consider $\{\bphi_k\}_{k=1}^K$ in our model; the probability weights for $\{\bphi_k\}_{k=K+1}^\infty$ in $\bpi_t$ will be added into $\pi_{tK}$. We note that a number of papers in topic modeling have put this approach into practice \cite{P10,W11}.

It has been shown \cite{P10} that when the truncation level $K$ is large, we may as well replace the distribution of $\bpi_1$ with $\bpi_1\sim Dir(\ga\bpi_0)$, where $\ga=1$, $\bpi_0=(1/K,\ldots,1/K)$. We also let $H=Dir(\be,\ldots,\be)$ as in the paper by Teh et al. (2005). We summarize our model,
\[
\begin{cases}
\bphi_1,\ldots,\bphi_K\iid Dir(\be,\ldots,\be),\\
\bpi_1 \sim Dir(\ga\bpi_0),\ \pi_{tk} = \pi_{t-1\, k}+Lap(\la),\\ 
\tbpi_t = \bpi_t + \bet\cdot\by_t,\\
\bth_{t,i}|(\al_t, \tbpi_t) \iid Dir(\al_t\tbpi_t), \\
z_{t,i,j}|\th_{t,i}\iid Cat(\bth_{t,i}),\ x_{t,i,j}\sim Cat(\bphi_{z_{t,i,j}}).
\end{cases}\addtag
\]
Here $t = 1,\ldots,T$, $i=1,\ldots,N_t$, $j=1, \ldots, J_{t,i}$ ($t\ge2$ for the second line). The last two lines above are derived as in Teh et al. (2005). We note that here $\bphi_k$ is the per-topic word distribution, $\bth_{t,i}$ is the per-document topic distribution, and $z_{t,i,j}$ is the actual topic for each word; they have the same meaning as in LDA.

We name our model a ``group-correlated temporal topic model'' (GCLDA). Here a ``group'' stands for all the documents within the same time period. We use the term "correlated" because the baseline topic distributions $\{\bpi_t\}_{t=1}^T$ for each period, controlling for $\{\by_t\}_{t=1}^T$, are endogenously correlated; meanwhile, the realized baseline topic distributions $\{\tbpi_t\}_{t=1}^T$ for each period are also correlated with the given exogenous process $\{\by_t\}_{t=1}^T$.
\subsection{Sampling the posterior: An MCMC Approach}
Direct estimation of the Bayesian posterior is often intractable since the closed-form expression, if it exists, can be difficult to integrate and thus, many approaches to approximate the posterior have been proposed. Monte Carlo methods, which draw a large number of samples from the posterior as its approximation, are particularly helpful. In this paper, we adopt the Markov chain Monte Carlo (MCMC) approach which constructs samples from a Markov chain and is asymptotically exact. Below we provide the Metropolis-within-Gibbs sampling approach tailored to our situation, which is a variant of the general MCMC approach. It only requires specifying the full conditionals of the unknown variables, which is covered below.

We consider sampling the following variables $\bZ=\{z_{t,i,j}\}_{j=1}^{J_{t,i}}\,_{i=1}^{N_t}\,_{t=1}^T$, $\{\al_t\}_{t=1}^T$, $\{\tbpi_t\}_{t=1}^T$, $\bet$, $\la$. We integrate out $\{\bth_{t,i}\}_{i=1}^{N_t}\,_{t=1}^T$ and $\{\bphi_k\}_{k=1}^K$ to speed up calculation. Following Griffiths and Steyvers (2004), conditioning on all other variables listed for sampling,
\begin{equation}
p(z_{t,i,j}=k|rest)\propto (C_{t,i,k}^{(-1)}+\al_t\tpi_{tk})\frac{C_{x_{t,i,k},k}^{(-1)}+\be}{C_{k}^{(-1)}+V\be}.
\end{equation}
Here $C_{t,i,k}^{(-1)}$ is the count of elements in $\bZ\bs\{z_{t,i,j}\}$ which belong to $d_{t,i}$ and has values equal to $k$; $C_{x_{t,i,k},k}^{(-1)}$ is the count of elements in $\bZ\bs\{z_{t,i,j}\}$ whose values are $k$ and corresponding words are $x_{t,i,j}$; $C_k^{(-1)}$ is the count of elements in $\bZ\bs\{z_{t,i,j}\}$ whose values are $k$. Also following Griffiths and Steyvers (2004), we have for $\al_t$ and $\tbpi_t$
\begin{align*}
&\qquad p(\al_t,\tbpi_t|rest)\propto p(\bZ|\al_t,\tbpi_t) p(\bpi_{1,\ldots,T}|\ga,\bpi_0)p(\al_t)\\
&\propto  \left[\frac{\Ga(\al_t)}{\prod_k\Ga(\al_t\tpi_{tk})}\right]^{N_t}\prod_{i=1}^{N_t}\frac{\prod_k\Ga(C_{t,i,k}+\al_t\tpi_{tk})}{\Ga(J_{t,i}+\al_t)} \\
&\times \exp \left[-\la\left( 1_{t<T}\cdot\|\pi_{t+1}-\pi_{t}\|_1+ 1_{t>1}\cdot\|\pi_{t}-\pi_{t-1}\|_1 \right)\right]\\
&\times p(\bpi_1|\ga\bpi_0)p(\al_t).\addtag
\end{align*}
Here the difference between $C_{t,i,k}$ and $C_{t,i,k}^{(-1)}$ is to replace $\bZ\bs\{z_{t,i,j}\}$ with $\bZ$. We also view $\bpi_t$, $\pi_{tk}$, etc. as functions of the other parameters; specifically, $\pi_{tk}=\tpi_{tk}-\bet_k\by_t$, where $\bet_k$ is the $k$-th row of $\bet$. This involves a transformation of variables; however, the related Jacobian determinant $\det(J)=1$, so (14) is still valid. For parameter $\bet$, we have
\begin{align*}
&\qquad p(\bet|rest)\propto \prod_{t=2}^Tp(\bpi_{t}|\bpi_{t-1},\la)\cdot p(\bet) \\
&\propto \exp\left(-\la\cdot\sum_{t=2}^T\|\pi_{t}-\pi_{t-1}\|_1\right)p(\bet).\addtag
\end{align*}
Finally, for parameter $\la$, we have
\begin{align*}
&\qquad p(\la|rest)\propto \prod_{t=2}^Tp(\bpi_{t}|\bpi_{t-1},\la)\cdot p(\la) \\
&\propto \la^{(T-1)K}\exp\left(-\la\cdot\sum_{t=2}^T\|\pi_{t}-\pi_{t-1}\|_1\right) p(\la). \addtag
\end{align*}
We note that (13) and (16) are full conditionals, and we can easily derive the full conditionals of $\al_t$, $\pi_{tk}$, and $\eta_k$ from (14) and (15). Since each $z_{t,i,j}|rest$ has a categorical distribution, and $\la|rest$ has a Gamma distribution with a conjugate prior, they can be updated with Gibbs updates. For $\al_t$, $\pi_{tk}$, and $\eta_k$, we replace a Gibbs update with a Metropolis update. Specifically, suppose we know $p(par|rest)$ up to a multiplicative constant, where $par$ is any length-$1$ parameter. We also assume $par=par^{(r)}$ at the $r$-th iteration. Then at the $(r+1)$-th iteration, 
\[
par_{new} \sim q(\cdot|par^{(r)}),
\]
\[
par^{(r+1)} = \left\{\begin{array}{ll}par_{new}\textrm{\ with\ prob.\ }P,\\ par^{(r)}\textrm{\ with\ prob.\ }1-P,\end{array}\right. \addtag
\] 
where $P=\min\{p(par_{new}|rest) /p(par^{(r)}|rest), 1\}$, and $q(\cdot|\cdot)$ is a known conditional probability distribution such that $q(x|y)=q(y|x)$. 

This completes our sampling and posterior inference. For the theoretical convergence properties of Metropolis-within-Gibbs samplers, the reader can refer to Robert and Casella (2004) and Roberts and Rosenthal (2006).

\section{Case Studies}
The proposed model, ``GCLDA,'' is demonstrated on two data sets: (1) online job advertisements from my.jobs from February to September in 2014, and (2) journalists' postings in 2014 in the ``Finance'' section in BusinessInsider.com, an American business and technology news website. Our algorithm has been implemented in Java, and we compare GCLDA with LDA, ToT, and STM.

\subsection{Experiment Settings}
We initialize the hyperparameters of LDA as follows: $\bal=(50/K,\ldots,50/K)$,  $\bbe=(0.01,\ldots,$ $0.01)$, according to a rule of thumb which has been carried out in Berry and Kogan (2010) and Sridhar (2015). For ToT, we use the same $\bal$ and $\bbe$ and linearly space the timestamps to make computation feasible. For GCLDA, we let $\ga=1$, $\bpi_0=(1/K,\ldots,1/K)$, $\al_t\iid \Ga(1,1)$, $p(\bet)\propto e^{-0.01\sum|\eta_k|}$, $\la\sim\Ga(1,1)$, $\be=0.01$. We carry out the Metropolis-within-Gibbs algorithm as described in Section 3.3 for GCLDA, and run $5\c000$ iterations of the Markov chain with $1\c000$ burn-in samples for GCLDA, LDA, and ToT; for LDA, we apply the collapsed Gibbs sampling as in Griffiths and Steyvers (2004). The number of topics is set to $K=50$ for both data sets. For STM, we apply the ``Spectral'' initialization \cite{R15b} together with other default settings in the R package \verb|stm|. We perform data cleaning, remove the stopwords, stem the documents, and keep most frequent $V$ words in each study. For the job advertisements, $V=2\c000$ and covers $96.2\%$ of all words with repetition, which means that the choice of words in job advertisements is quite narrow; for the journalists' postings, $V=3\c000$ and covers $93.9\%$ of all words with repetition. 

We use perplexity to compare the difference of the prediction power between LDA, ToT, and GCLDA. The perplexity for $N_{test}$ held-out documents given the training data $D$ is defined as 
\begin{equation}
perp = \exp\left\{-\dfrac{\sum_{i=1}^{N_{test}}\log p(d_{test,i}|D)}{\sum_{i=1}^{N_{test}}n_{test,i}}\right\}
\end{equation}
where $d_{test,i}$ represents the $i$-th held-out document, and $n_{test,i}$ is the number of words in $d_{test,i}$. We expect the perplexity to be small when a model performs well, since this means that under the estimated model, the probability of a word in the testing documents being written {\it a priori} is large. We apply the ``Left-to-right'' algorithm \cite{W09} and apply point estimates for ``$\Phi$'' and ``$\al\b{m}$'' using the training data, as suggested in Section 3 in the same paper. 

\subsection{My.jobs: Online Job Advertisements}
The number of online job advertisements on my.jobs from February to September in 2014 amounts to $17\c147\c357$ in total, and the number of advertisements each day varies greatly. Therefore, we gather a stratified sample of $44\c660$ advertisements with a roughly equal number of samples for each day, so that we have sampled $0.26\%$ of all the documents in total. The training data set consists of $40\c449$ advertisements, and the testing data set consists of $4\c211$ advertisements ($9.4\%$ of the sample). For the exogenous variable $\{\by_t\}_{t=1}^T$, we use the standardized Consumer Price Index from February to September in 2014, so that $p=1$, and $T=8$.

Figure 1 implies that GCLDA better predicts the words in the new documents in terms of perplexity. This is due to the fact that the introduction of both endogenous and exogenous processes allows us to make more accurate inference on the topic distributions of the documents in a given period of time. The standard errors for the perplexity in each period are also shown; we can observe that the difference is quite significant.
\begin{figure}[H]
	\centering
	\includegraphics[scale=1]{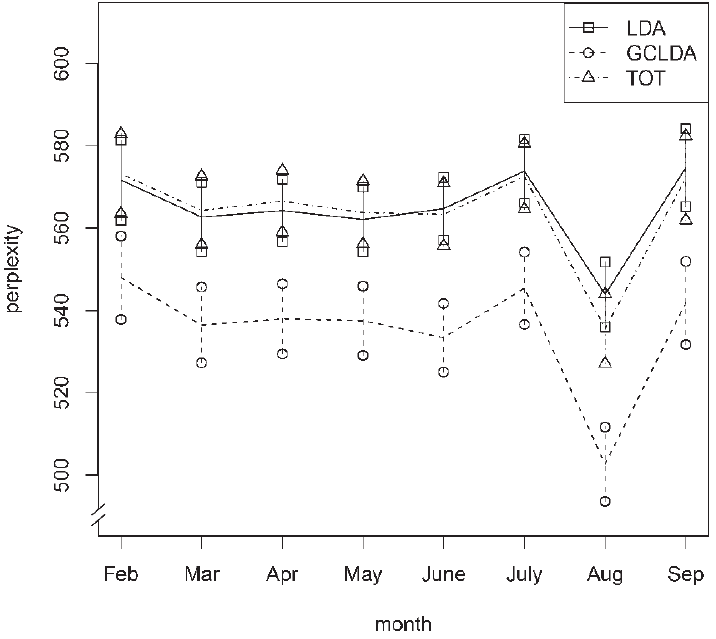}
	\caption{Perplexity results for the job advertisements from February to September in 2014.}
\end{figure}
The $20$ most common topics are presented in Figure 2. The only axis, the x-axis,  represents the degree of correlation $\rho=\eta/\pi$ for all topics, i.e. the percent change in the topic proportion given one unit change in the exogenous covariate. Here $\pi$ and $\eta$ denote the related component of $\sum \bpi_t/T$ and $\bet$ for each topic. Table 1 lists the highest probability words sorted by their probabilities from high to low inside the five topics with highest $\rho$ in Figure 2.
\begin{figure}[H]
	\centering
	\includegraphics[scale=0.64]{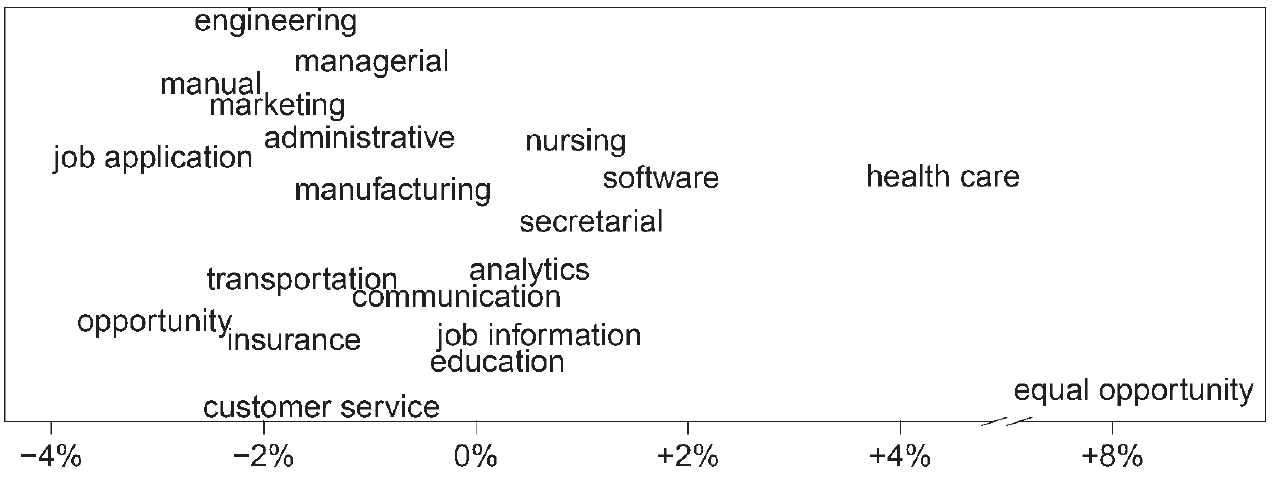}
	\caption{The 20 most common topics and their $\rho$ from GCLDA for the job advertisements.}
\end{figure}
\begin{table}[H]
	\footnotesize
	\centering
	\begin{tabular}{|l|}
		\hline {\it equal opportunity}: employment status disabled veteran equal\\ 
		\hline {\it health care}: health care medical service provide center hospital \\ 
		\hline {\it software}: development experience software design application\\ 
		\hline {\it secretarial}: management operations ensure training perform\\ 
		\hline {\it nursing}: care nursing patient required clinical practice medical \\
		\hline
	\end{tabular}
	\caption{The highest probability words inside the five topics with highest $\rho$ in Figure 2.}
\end{table}
A number of facts can be inferred from Figure 2. The topics with positive $\rho$ are those that have a positive correlation with the growth of the CPI in 2014. We can observe that two of them are supported by the U.S. government spending, namely ``equal opportunity'' and ``health care,'' the latter of which is probably related to the Affordable Care Act programs. This suggests that there is a causal relationship between the increase in government spending and the increase in the number of jobs in these categories, and the former was also an underlying factor in the growth of the CPI in 2014. We also observe that ``software'' and ``secretarial'' were moving in the same direction of CPI, while some traditional higher-paid job categories, such as engineering and marketing, were not. This partly agrees with some news articles in 2014 in that while the labor market was recovering, there was relatively lower growth in traditional higher-paid job categories\linebreak \cite{L14,C14}.

Therefore, we posit that GCLDA could be helpful in identifying topics in temporal documents which are closely related to an exogenous process. The changes in the topic proportions may be caused by the exogenous process, or its underlying factors, as is illustrated by this example, where more demand of goods increases the CPI and creates more jobs in certain categories. The other way around is also possible; changes in an exogenous process are caused by certain kinds of news, as is demonstrated in the next example. Such relationships require a more case-specific examination.

We also compare our method with STM. Below is the STM counterpart of Figure 2. We can observe that the topics and correlation scores from GCLDA seem to be more time-related and tend to be more informative of the labor market during the period.
\begin{figure}[H]
	\centering
	\includegraphics[scale=0.66]{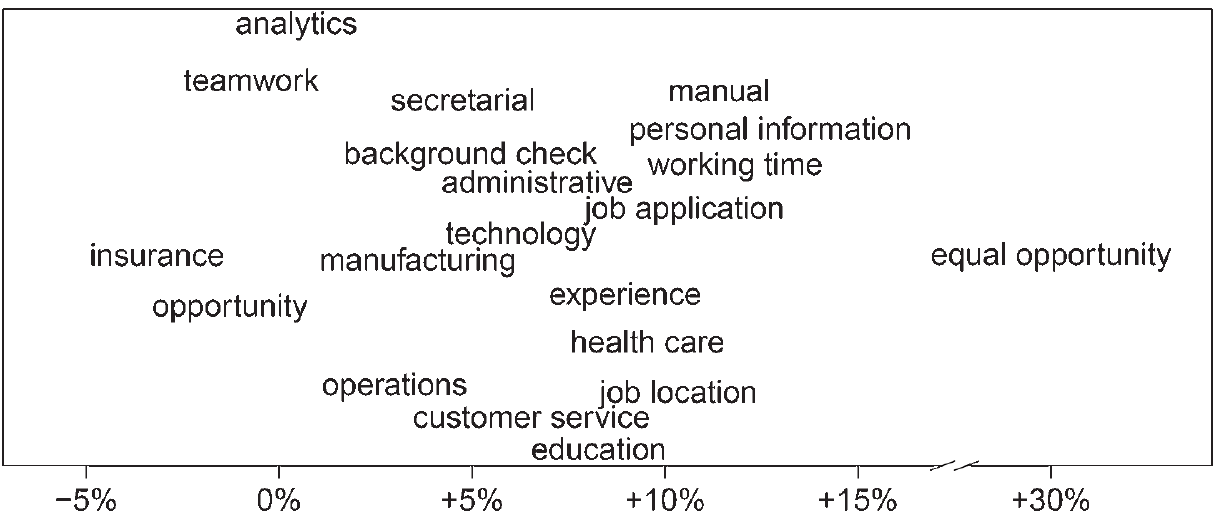}
	\caption{The 20 most common topics and their $\ga$ from STM for the job advertisements.}
\end{figure}

\subsection{BusinessInsider.com: Financial News Articles}
We consider all contributions in the ``Finance'' section of BusinessInsider.com on all trading days in 2014. There are $15\c659$ articles in total, which are divided into a training data set containing $12\c527$ articles and a testing data set containing $3\c132$ articles ($20\%$ of all articles). We increase the proportion of testing documents and let $T=252$ (all trading days) in order to create a more challenging scenario for GCLDA. We apply the daily price of the Chicago Board Options Exchange Market Volatility Index (VIX) as the exogenous process, measuring the volatility of the U.S. financial market. The other settings are the same as those in Section 4.2. We provide an analysis of the perplexity of LDA, GCLDA, and ToT in Figure 4. The lines are smoothed by LOESS with a span of $0.2$, as there are large fluctuations in perplexity from day to day.
\begin{figure}[H]
	\centering
	\includegraphics[scale=0.9]{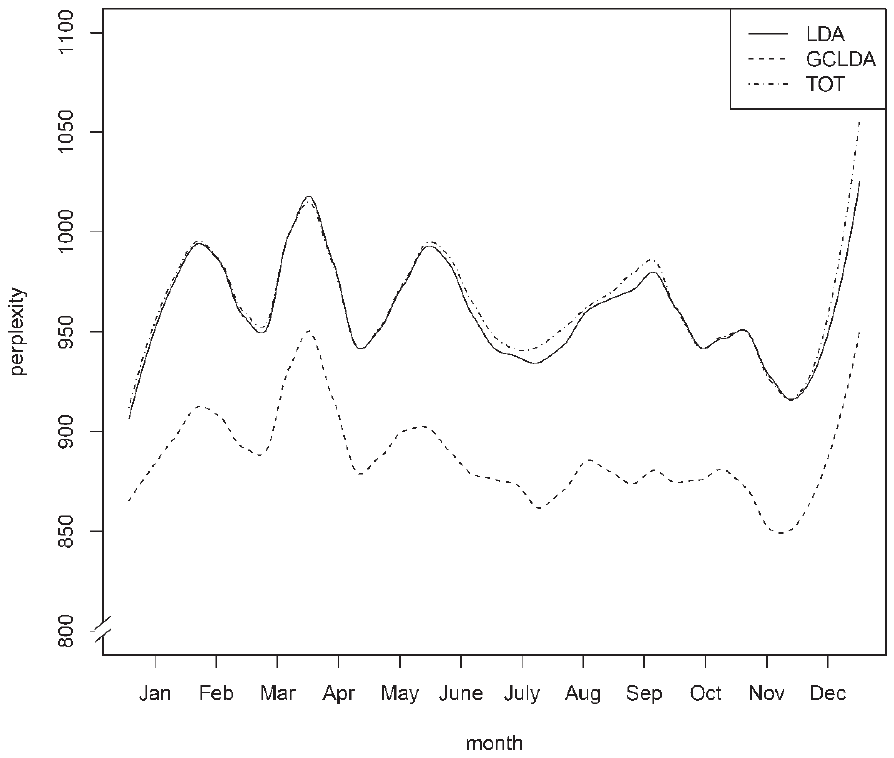}
	\caption{Perplexity results for contributions in the ``Finance'' section in BusinessInsider.com in 2014.}
\end{figure}
Again we observe that GCLDA generates a lower perplexity for the testing documents over time, therefore improving the fitting of the topic model. Also, the perplexities obtained from LDA and ToT are largely the same. 

From Figure 5 and Table 2 below, the topics that are strongly positively correlated with the VIX are generally short-term news, such as stock market news and announcements from central banks, as in the topics ``FED'' (the Federal Reserve), ``revenue,'' and ``stock market,'' which are indeed closely related to changes in the stock market. From our analysis, the drop in oil price and the instability in Russia and Ukraine were also major causes of fluctuations in the stock market in 2014. On the other hand, we observe that news about longer-term economic trends is not positively correlated with the VIX, such as ``companies'' and ``labor market.'' These are generally consistent with our understanding of the stock market. 

In this example, we demonstrate an application of GCLDA for finding documents that are major contributors to changes in an exogenous process during a period of time. We can easily estimate the topic distribution for each document, and therefore, we can select the news articles that are mostly related to the stock market. Such a direction could possibly evoke future research. We also note that in this example, the causal relationship between the topic distribution and the stock market is bidirectional; news can change the stock market, and {\it vice versa}.
\begin{figure}[H]
	\centering
	\includegraphics[scale=0.65]{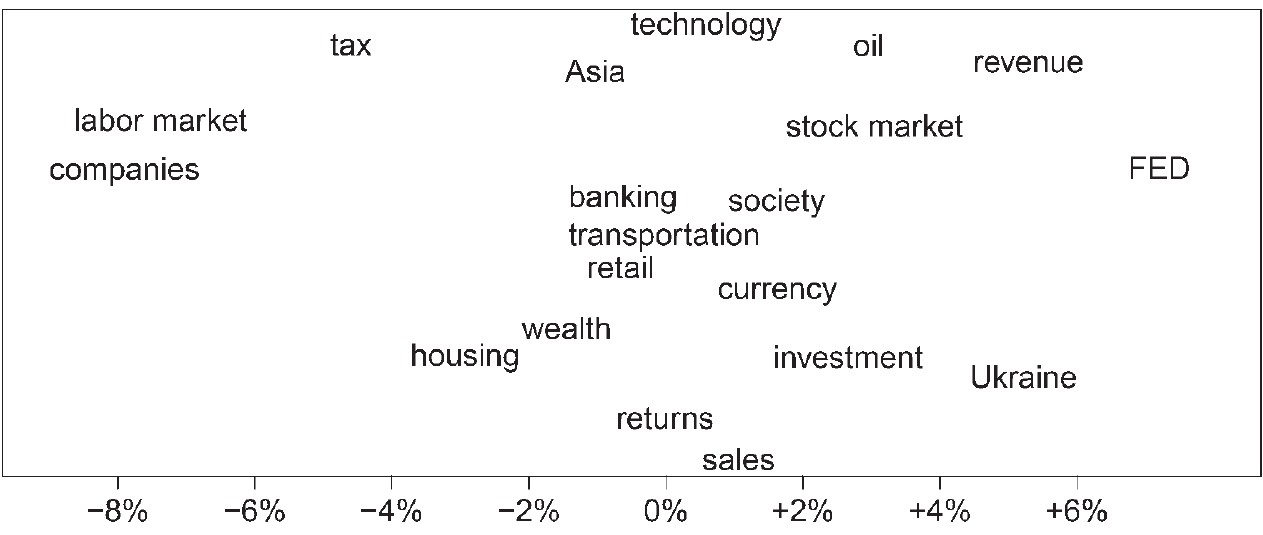}
	\caption{The 20 most common topics and their $\rho$ from GCLDA for the financial articles.}
\end{figure}
\begin{table}[H]
	\footnotesize
	\centering
	\begin{tabular}{|l|}
		\hline
		{\it FED}: rate FED inflation policies market federal expected\\ \hline
		{\it revenue}: quarter billion year revenue million earnings share\\  \hline
		{\it Ukraine}: Russia Ukraine Moscow gas country president\\  \hline
		{\it stock market}: market trade stock week day close morning\\  \hline
		{\it energy}: oil price energies gas production crude supplies\\	\hline
	\end{tabular}
	\caption{The highest probability words inside the five topics with highest $\rho$ in Figure 5.}
\end{table}
We also compare our method with STM, with Figure 6 being the STM counterpart of Figure 5. Again we observe that the topics and correlation scores from GCLDA seem to be more time-related and tend to be more informative of the stock market during the period. These findings assert our view that GCLDA improves the structure of the topic model and makes it more time-dependent. 
\begin{figure}[H]
	\centering
	\includegraphics[scale=0.64]{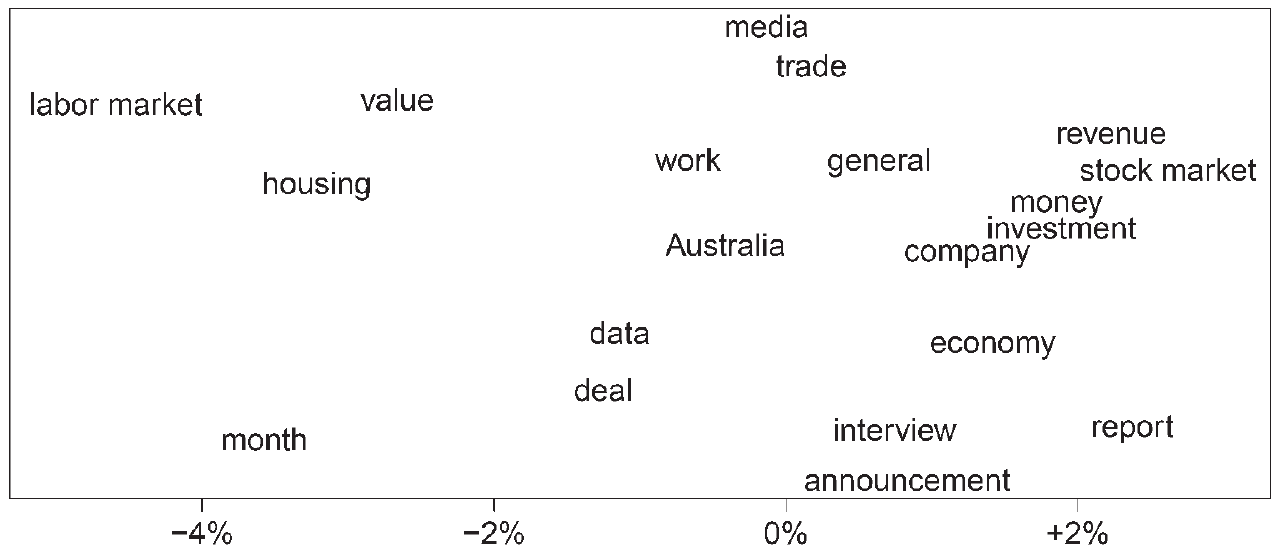}
	\caption{The 20 most common topics and their $\ga$ from STM for the financial articles.}
\end{figure}
\section{Conclusion}
We have developed a temporal topic model which analyzes time-stamped text documents with known exogenous processes. Our new model, GCLDA, takes both endogenous and exogenous processes into account, and applies Markov chain Monte Carlo sampling for calibration. We have demonstrated that this model better fits temporal documents in terms of perplexity, and extracts well information from job advertisements and financial news articles. We suggest that a possible direction for the future could be analyzing the contents of temporal documents so that they could predict the trends of related exogenous processes.
\section*{Acknowledgments}
This research was conducted in collaboration with the Workforce Science Project of the Searle Center for Law, Regulation and Economic Growth at Northwestern University. We are indebted to Deborah Weiss, Director, Workforce Science Project, for introducing us to the subject of workforce and providing guidance. We are also very grateful for the help and data from DirectEmployers Association.
\nocite{B10,F73,G04,R04,R06,S15,T05} 
\bibliography{b1}
\bibliographystyle{aaai}
\end{document}